\documentclass[conference]{IEEEtran}
\IEEEoverridecommandlockouts
\usepackage{cite}
\usepackage{amsmath,amssymb,amsfonts,multirow}
\usepackage{graphicx}
\usepackage[table,xcdraw]{xcolor}
\usepackage{textcomp}
\usepackage{xcolor}
\usepackage{ctable} 
\usepackage{subcaption}
\usepackage{soul}
\usepackage[ruled,vlined,linesnumbered]{algorithm2e}
\usepackage{algpseudocode}
\usepackage[nodisplayskipstretch]{setspace}
\usepackage{authblk}

\def\BibTeX{{\rm B\kern-.05em{\sc i\kern-.025em b}\kern-.08em
 T\kern-.1667em\lower.7ex\hbox{E}\kern-.125emX}}
\begin{document}

\title{A Safe Genetic Algorithm Approach for Energy Efficient Federated Learning\\ in Wireless Communication Networks}

\author{Lina Magoula*, Nikolaos Koursioumpas*, Alexandros-Ioannis Thanopoulos*,\\Theodora Panagea*, Nikolaos Petropouleas*,\\ 
M. A. Gutierrez-Estevez**, Ramin Khalili**
\\
* \emph{Dept. of Informatics and Telecommunications, National and Kapodistrian University of Athens, Greece} 
\\
** \emph{Munich Research Center, Huawei Technologies Duesseldorf GmbH, Munich, Germany}
\\

\{lina-magoula*, nkoursioubas*, gthanopoulos*, dpanagea*, nipet*\}@di.uoa.gr \\

\{m.gutierrez.estevez**, ramin.khalili**\}@huawei.com}

\maketitle

\begin{abstract}
  Federated Learning (FL) has emerged as a decentralized technique, where contrary to traditional centralized approaches, devices perform a model training in a collaborative manner, while preserving data privacy. Despite the existing efforts made in FL, its environmental impact is still under investigation, since several critical challenges regarding its applicability to wireless networks have been identified. Towards mitigating the carbon footprint of FL, the current work proposes a Genetic Algorithm (GA) approach, targeting the minimization of both
  the overall energy consumption of an FL process and any unnecessary resource utilization, by orchestrating the computational and communication resources of the involved devices, while guaranteeing a certain FL model performance target. A penalty function is introduced in the offline phase of the GA that penalizes the strategies that violate the constraints of the environment, ensuring a safe GA process. Evaluation results show the effectiveness of the proposed scheme compared to two state-of-the-art baseline solutions, achieving a decrease of up to 83\% in the total energy consumption.
\end{abstract}


\section{Introduction}\label{Intro}
The Information and Communication Technology (ICT) industry represents an important energy consumer, utilizing 4\% of the world’s
electricity \cite{LANGE2020106760}. The ICT industry is expected to reach 10-20\% of the world’s
electricity by 2030 \cite{global5g},\cite{challe6010117}. Many ambitious climate activities are already taking place by European Commission across the ICT industry, in line with the GSMA’s commitment towards net-zero carbon emissions by 2050 \cite{GSMA_1, EU_Regulations}.  

In future wireless networks, the use of Artificial Intelligence (AI) as a key enabler has been recognized at European and global level \cite{kaloxylos_alexandros_2020_4299895}. Centralized AI approaches require a vast amount of data transfer to centrally located and energy hungry data centers, raising concerns with regard to both data privacy and energy consumption \cite{Savazzi9807354}. Alternative solutions such as Federated Learning (FL) have surfaced, where devices with typically low power profiles perform a model training in a collaborative manner, exploiting locally stored datasets and avoiding any raw data transmissions. The 3rd Generation Partnership Project's (3GPP) in Rel. 18, tries to adopt FL in Network Data Analytics Function (NWDAF) \cite{3GPP-FL}, with the introduction of the Model Training Logical Function (MTLF) in Rel. 17 that is responsible of training Machine Learning (ML) models and exposing new training services \cite{ETSI-MTLF}. 

Several critical energy related challenges have been raised related to the application of FL to wireless networks \cite{Niknam9141214}. Firstly, the model training process requires from the computing devices constant model update transmissions. Such model updates could consist of thousands or even millions of parameters comprising complex neural networks. This implies a significant communication overhead, making the uplink transmissions particularly challenging. Secondly, ML model architectures require complex calculations, tightly related to the size of the ML model, resulting in even more challenges in terms of the energy aspect, since significant computational work will take place on device level during training. Another critical challenge of major importance includes computation and data heterogeneity across devices, in conjunction with strict ML model performance requirements. More specifically, critical services and applications often require stringent ML model performance \cite{5GCroCo_D2_1}, which is directly translated to higher training times and as a result to an increase in the overall energy consumption. Such deterioration could be accelerated in case of devices differing in both available resources and statistical data distribution. 

Taking into consideration the aforementioned challenges, it becomes apparent that new solutions are required in order to mitigate the environmental impact of FL in wireless communication networks, while guarantying a certain ML model performance. A number of state-of-the-art works focus on energy efficient FL, tackling the problem in terms of resource orchestration, computation offloading, as well as load balancing strategies (e.g. \cite{9473734Ren,8488502Cao,9145118Zeng,9545925Wang,Yang13037554,9384231Zhan,9475121Mo}).

Contrary to the existing efforts, the current work proposes a Genetic Algorithm (GA) approach, targeting the minimization of both the overall energy consumption of an FL process and any unnecessary resource utilization, by orchestrating the computational and communication resources of the involved devices, while guaranteeing a certain FL model performance target. The proposed solution considers the FL model’s complexity (proposed only in \cite{9475121Mo,9384231Zhan}), as well as device and data heterogeneity, as part of the overall energy consumption. Towards a safe decision making, a penalty function is introduced in the offline phase of the GA that penalizes the strategies that violate the constraints of the environment, ensuring a safe GA process. Finally, considering the
potentially prohibited energy cost that is required for repetitive
FL executions, while exploring for optimal solutions, this is the only work that proposes a simulated and computationally cost effective FL environment that emulates a real FL process, solely in the offline phase. 

The rest of the paper is organized as follows. Section \ref{system_model} provides the system model. Section \ref{problem_formulation} provides the problem formulation. Section \ref{proposed_solution} describes the proposed solution that is evaluated in Section \ref{performance_evaluation} using the simulation setup of Section \ref{simulation_setup}. Finally, section \ref{conclusions} concludes the paper.
\vspace{-2pt}
\section{System Model} \label{system_model}
\vspace{-2pt}

This section provides the system model of an AI-enabled  wireless communication network using FL. FL enables collaborative decentralized training of a Deep Neural Network  (DNN), across multiple and heterogeneous network devices -  acting as workers-, without exchanging local data samples. The  set of workers, denoted by $\mathcal{K}$, is orchestrated by a coordinator  node. 

An FL process is comprised of a number of rounds. Each round is named as \textit{global} iteration and is denoted by $n \in \mathbb{N}$. Let $\textbf{w}_n$ be a vector containing the model parameters at \textit{global} iteration $n$. In each iteration $n$, the coordinator node distributes the $\textbf{w}_n$ of a global DNN model of size $m$, in bits, to the worker nodes that are involved in the FL process. Let $\alpha$ denote the complexity associated with the global model and measured in Floating Point Operations (FLOPs). Each worker $k \in \mathcal{K}$, after successfully receiving the vector with the global parameters $\textbf{w}_n$, performs a local training using its locally stored data samples $\mathcal{D}_k$. 

\textbf{Device Computational Capability}: Let $f_{k,n}$ denote the available computational capacity (i.e. Central Processing Unit (CPU) speed) of worker $k$ to execute a local training at the $n^{th}$ \textit{global} iteration (it is assumed that the coordinator node has enough computational capacity to execute any computational task). Based on $f_{k,n}$, the worker $k$ can complete a certain number of FLOPs per cycle, denoted by $c_{k,n}$. The CPU of each worker $k$ has also an effective switched capacitance, denoted by $\varsigma_k$, which depends on the hardware architecture. The local training of a worker is considered complete when a pre-selected performance target $\eta$ is reached at each worker. The performance target $\eta$ could refer to error metrics (e.g. mean squared error), accuracy, etc. Let $I_{k,n}$ be the total number of \textit{local} iterations required in order for the locally trained  model of worker $k$ to reach the pre-selected performance target $\eta$ at \textit{global} iteration $n$. By $\tau_{k,n}$ is denoted the time required by $k$ to complete a local training process.

\textbf{Device Communication Capability: } After $I_{k,n}$ \textit{local} iterations, each $k$ produces a locally trained model with parameters $\textbf{w}_{k,n}$ that needs to be transmitted to the coordinator. The communication channel between worker $k$ and the coordinator at \textit{global} iteration $n$ is modeled as a flat-fading channel with Gaussian noise power density $N_0$ and channel gain $g_k$, where the fading is assumed constant during the transmission of the model. Also, let $b_{k,n}$ and $p_{k,n}$ be the assigned bandwidth and transmission power to $k$ (it is assumed that enough bandwidth has been assigned to $k$ to transmit its model updates). $r_{k,n}$ and $tr_{k,n}$ are the achievable data rate and the required time to upload $\textbf{w}_{k,n}$ to the coordinator, respectively (see Section \ref{problem_formulation} for more details).

\textbf{FL Rounds: } The FL process is realized in a synchronized manner. A \textit{global} iteration is finished when the coordinator receives updates from all workers or when a preselected time threshold $\mathsf{H}$ is reached. Workers therefore should complete their computational and communication tasks within this time threshold (Eq. (\ref{complete_version_constraint_total_latency})): Updates from those who have not met the time threshold are considered invalid and are not used. The coordinator node produces an updated global model $\textbf{w}_{n+1}$, using all the received model updates, and transmits it back to the workers to start the next \textit{global} iteration ($n+1$). This procedure is repeated until the global model reaches a pre-selected performance target $\epsilon_0$. Table \ref{table_notations} summarizes the notations.

\begin{table}[ht!]
\centering
\renewcommand{\arraystretch}{0.98}
\resizebox{255pt}{!}{%
\begin{tabular}{|m{0.035\textwidth}|m{0.46\textwidth}|}
\hline
$\mathcal{K}$ & Set of worker nodes \\ \hline
$n$ & Index of the \textit{global} iteration (FL round) \\ \hline
$N_0$ & White Gaussian noise power spectral density\\ \hline
$g_k$ & Gain of the wireless channel the worker $k$ has access to\\ \hline
$f_{k,n}$ & Available computational capacity of worker $k$ at the $n^{th}$ \textit{global} iteration\\ \hline
$b_{k,n}$ & Bandwidth assigned to worker $k$ at the $n^{th}$ \textit{global} iteration\\ \hline
$p_{k,n}$ & Transmission power of worker $k$ at the $n^{th}$ \textit{global} iteration \\ \hline
$r_{k,n}$ & Achievable transmission data rate of worker $k$ at the $n^{th}$ \textit{global} iteration\\ \hline
$\mathcal{D}_k$ & Local data samples of worker $k$ \\ \hline
$c_{k,n}$ & Total number of FLOPs per cycle that the worker $k$ can complete at the $n^{th}$ \textit{global} iteration\\ \hline
$\varsigma_k$ & Effective switched capacitance of worker $k$ \\ \hline
$\textbf{w}_n$ & Global FL model produced at the $n^{th}$ \textit{global} iteration \\ \hline
$\alpha$ & Complexity of the global FL model in terms of total number of FLOPs \\ \hline
$m$ & Size of global FL model in bits \\ \hline
$I_{k,n}$ & Number of \textit{local} iterations required to reach $\eta$ at the worker $k$ at the $n^{th}$ \textit{global} iteration\\ \hline
$\tau_{k,n}$ & The time required by a worker $k$ to complete a local training process at the $n^{th}$ \textit{global} iteration\\ \hline 
$tr_{k,n}$ & The time required by worker $k$ to transmit its model updates at the $n^{th}$ \textit{global} iteration\\ \hline
$\textbf{w}_{k,n}$ & Model parameters produced by worker $k$ at the $n^{th}$ \textit{global} iteration \\ \hline
$\eta$ & Pre-selected performance target imposed to the training process of all workers\\ \hline
$\epsilon_0$ & Pre-selected performance target imposed to the global FL model\\ \hline
\end{tabular}
}
\caption{Notation Table} \label{table_notations} 
\end{table}
\section{Problem Formulation} \label{problem_formulation}
\vspace{-6pt}

The objective is to achieve energy efficiency in the system, i.e minimizing both the overall energy consumption of the workers and any unnecessary
resource utilization, while guarantying a certain global model performance $\epsilon_0$. In each \textit{global} iteration $n$, all workers consume a specific amount of energy from the power grid, denoted by $E^{gr}_{n}$, in order to complete their tasks. Two types of tasks are considered in an FL process, the computation tasks, referring to the local training processes, and the communication ones, related to the transmission of the model parameters. In case a worker does not complete its tasks in the given time threshold $\mathsf{H}$, its allocated resources are considered wasted (unnecessary
resource utilization), since there is no contribution to the \textit{global} model update by this worker. 


The amount of computation and transmission energy required by a worker $k$ during \textit{global} iteration $n$ are denoted by $E_{k,n}^C$ and $E_{k,n}^T$, respectively, given by: 
\vspace{-7pt}
\begin{equation}  \label{computation_energy}
  E_{k,n}^C = \frac{\varsigma_{k} \cdot I_{k,n} \cdot \alpha \cdot \mathcal{D}_k \cdot f_{k,n}^2}{c_{k,n}},
\end{equation}
\vspace{-4pt}
and
\vspace{-4pt}
\begin{equation} \label{transmission_energy}
  E_{k,n}^T = \frac{m \cdot p_{k,n}}{b_{k,n} \cdot \log_2\left(1+\frac{g_k \cdot p_{k,n}}{N_0 \cdot b_{k,n}}\right)}.
\end{equation}

Since the number of \textit{global} iterations required for the model to converge to $\epsilon_0$ is not known a priori, 
a greedy approach is followed where, at each iteration $n$, we attempt to minimize the total energy consumption required by all workers to complete their computational and communication tasks, while a certain local accuracy $\eta$ is achieved at each worker. This greedy approach is repeated until the global model reaches the pre-selected performance target $\epsilon_0$. Although an FL process lacks theoretical convergence guarantees \cite{Li2020On}, for the task that we study and the data distribution considered, the convergence to $\epsilon_0$ is always achieved (Section \ref{performance_evaluation}). The parameters of the optimization are the computational capacity $f_{k,n}\in \mathbb{R}_+$ and transmission power $p_{k,n} \in \mathbb{R}_+$ of each worker $k$ at each iteration $n$ of the FL process. The objective function is defined as:
\vspace{-12pt}
\begin{eqnarray} \label{objective_function}
  \min_{\textbf{f}_n, \textbf{p}_n} E^{gr}_{n} = \sum_{k=1}^{\mathcal{K}} (E_{k,n}^C\cdot \Omega_{k,n} + E_{k,n}^T)
  \vspace{-4pt}
\end{eqnarray}
\vspace{-10pt}
\begin{eqnarray} \label{complete_version_constraint_total_latency}
 \textrm{s.t.}&\tau_{k,n} + tr_{k,n} < \mathsf{H}, \forall k \in \mathcal{K} 
\end{eqnarray}
\vspace{-18pt}
\begin{eqnarray} && \label{complete_version_constraint_per_device_capacity}
  0\leq f_{k,n} \leq f_k^{max}, \forall k \in \mathcal{K} 
\end{eqnarray}
\vspace{-18pt}
\begin{eqnarray} && \label{complete_version_constraint_per_device_power}
  0\leq p_{k,n} \leq p_k^{max} ,  \forall k \in \mathcal{K}
\end{eqnarray}
\vspace{-18pt}
\begin{eqnarray} && \label{complete_version_constraint_idle}
 \sum_{k=1}^{\mathcal{K}} f_{k,n} > 0,
\end{eqnarray}
\vspace{-9pt}
where:
\begin{equation} \label{computation_time}
  \tau_{k,n}= \frac{I_{k,n} \cdot \alpha \cdot M_k}{c_{k,n} \cdot f_{k,n}}, 
\end{equation}
\begin{equation} \label{transmission_time}
  tr_{k,n}= \frac{m}{r_{k,n}},
\end{equation}
\begin{equation}
  r_{k,n} = {b_{k,n} \cdot \log_2\left(1 + \frac{g_k \cdot p_{k,n}}{b_{k,n} \cdot N_0}\right)}.\label{data_rate}
\end{equation}
More precisely, $\textbf{f}_n = [f_{1,n}, f_{2,n},...,f_{\mathcal{K},n}]^T$, $\textbf{p}_n = [p_{1,n}, p_{2,n},...,p_{\mathcal{K},n}]^T$, and $\Omega_{k,n}$ (Eq. \ref{indicator_function_omega}) is an indicator function ensuring that there will be no local training at  worker $k$ at the iteration $n$, in case of zero allocated transmission power:
\vspace{-4pt}
\begin{equation} \label{indicator_function_omega}
  \Omega_{k,n} = 
   \left\{
  \begin{array}{ll}
  1, & p_{k,n} > 0\\
  0, & otherwise\\
  \end{array}
  \right.
  \vspace{-4pt}
\end{equation}

 Constraint (\ref{complete_version_constraint_total_latency}) ensures the synchronization of the FL process, by upper bounding the total time required by each worker to complete a computation and transmission task at the time threshold $\mathsf{H}$. Constraints (\ref{complete_version_constraint_per_device_capacity}) and (\ref{complete_version_constraint_per_device_power}) ensure that the computational capacity along with the transmission power of a worker are within its maximum capabilities, denoted by $f_k^{max}$ and $p_k^{max}$, respectively. Finally, constraint (\ref{complete_version_constraint_idle}) ensures that at least one worker should be involved in the FL process. 
\vspace{-3pt}
\section{Proposed Genetic Algorithm Solution} \label{proposed_solution}
In this section, a safe GA meta-heuristic approach is proposed. GA provides a feasible solution to strategically perform a global search by means of many local searches, generating high-quality solutions relying on biologically inspired operations, such as parent selection, crossover and mutation \cite{Mirjalili2019}. A GA in each generation, constructs a population of \emph{chromosomes}, which is a set of candidate solutions to the optimization problem. The target of the GA is to provide higher quality solutions over the generations using as criterion the fitness score of each chromosome, which represents the target metric of the optimization problem. The main definitions of the GA solution are provided below:\\
\textbf{Generation: } A \textit{global} iteration $n$ of an FL process.\\
\textbf{Gene: } A resource assignment to worker $k$ at \textit{global} iteration $n$, i.e. $[f_{k,n}, p_{k,n}]$, that is bounded according to constraints (\ref{complete_version_constraint_per_device_capacity}) and (\ref{complete_version_constraint_per_device_power}).\\
\textbf{Chromosome: } A vector consisting of genes equal to the number of workers, representing a candidate solution to the optimization problem.\\
\textbf{Elites: } The number of best-scored chromosomes in a generation that are included unaltered in the next generation.\\
\textbf{Population: } A fixed-size set of chromosomes.\\
\textbf{Fitness Function: } A function that serves as a score for each chromosome and is formulated based on the objective function (\ref{objective_function}), in conjunction with constraints (\ref{complete_version_constraint_total_latency}) and (\ref{complete_version_constraint_idle}). Specifically:
\vspace{-4pt}
\begin{equation}\label{sac_reward}
FF_{n} = - [\sum_{k=1}^{\mathcal{K}}(E^C_{k,n}\cdot \Omega_{k,n} + E^T_{k,n}) + v_{n}]
\vspace{-4pt}
\end{equation}

The first part of Eq. (\ref{sac_reward}) is the objective function of our problem formulation. The second part $v_{n}$ is a penalty term defined to guarantee a safe GA process \cite{2886795Garc}. As a result, the chromosomes that violate the constraints, wasting resources, will be penalized. We define this penalty as follows:
\vspace{-3pt}
\begin{equation} \label{penalty}
  v_{n} = \sum_{k=1}^{\mathcal{K}}{(E_{k,n}^W + \mu_1 \cdot P^{(1)}_{k,n})} + \mu_2 \cdot P^{(2)}_{n} + P^{(3)}_{n}
  \vspace{-3pt}
\end{equation}
where $E_{k,n}^W$ is the wasted energy consumption of worker $k$, $\mu_1$ and $\mu_2$ are constant penalty weights of each constraint violation, and $P^{(1)}_{k,n}$, $P^{(2)}_{n}$ are two indicator functions, related to constraints (\ref{complete_version_constraint_total_latency}) and (\ref{complete_version_constraint_idle}). Considering also the total energy consumption of a complete FL process, apart from the energy consumed in individual FL rounds, an indicator function $P^{(3)}_{n}$ is introduced as part of the penalty function. This function penalizes the strategies that result in higher total energy consumption for a complete FL process $l$, denoted by $E_{FL}^{(l)}$, compared to the previous one ($E_{FL}^{(l-1)}$):
\vspace{-2pt}
\begin{equation} \label{device_constraint_violation_indicator_1}
  P_{k,n}^{(1)} = 
  \left\{
  \begin{array}{ll}
  0, & \tau_{k,n}+tr_{k,n}-\mathsf{H}<0 \\
  1, & otherwise,\\
  \end{array}
  \right.
  \vspace{-6pt}
\end{equation}

\begin{equation} \label{device_constraint_violation_indicator_2}
  P_{n}^{(2)} = 
  \left\{
  \begin{array}{ll}
  0, & \sum_{k=1}^{\mathcal{K}} f_{k,n} > 0\\
  1, & otherwise,\\ 
  \end{array}
  \right.
\end{equation}
\vspace{-4pt}
\begin{equation} \label{total_energy_indicator_3}
  P_{n}^{(3)} = 
\left\{
  \begin{array}{ll}
  -\frac{E_{FL}^{(l-1)}-E_{FL}^{(l)}}{E_{FL}^{(l)}}, & E_{FL}^{(l)} < E_{FL}^{(l-1)}\\
  \frac{E_{FL}^{(l)}-E_{FL}^{(l-1)}}{E_{FL}^{(l-1)}}, & otherwise\\
  \end{array}
  \right.
\vspace{-4pt}
\end{equation}

The GA starts with a randomly selected population in the first generation and targets to maximize (\ref{sac_reward}) across generations, which results in minimizing the total energy consumption and the wasted resources, with respect to the constraints of the system.\\
\textbf{GA Operations: } In each generation a Roulette Wheel method is used for parent selection, along with a uniform crossover and mutation \cite{Thomas1996} operations, with pre-selected probabilities, named as crossover and mutation rate. Apart from the standard operations, the GA is enhanced with a hybrid operation \cite{Macedo16} that combines the Triggered Hyper-mutation (TH) and Fixed Memory (FM). When a change in the system is detected, the TH increases the mutation rate for a certain number of generations, while the FM that uses a fixed-length memory, replaces the worst-scored chromosome from the population with the best individual from the memory. This hybrid operation is triggered when the distance of the fitness score from the best-scoring chromosomes of two consecutive generations is higher than a pre-selected percentage threshold $\mathsf{D}$.\\
\textbf{Early Stopping: } A termination criterion is applied to GA that interrupts the searching procedure, when the best solution during the evolution process does not improve for a certain number of generations.\\
\textbf{Offline and Online Phase: } During the offline phase of the GA, in each generation, an FL \textit{global} iteration is executed as many times as the population size, since different candidate solutions are tested. Note that, embedding periodical executions of FL rounds, is of high complexity and time consuming, but most importantly, contrary to the objective of the current work; it requires a significant amount of energy consumption. In order to overcome this issue and as it will be described in Section \ref{simulation_setup}, a simulated and significantly simpler FL environment was designed for the GA's offline phase, emulating a real FL process for a given performance target. Finally, the online phase will use real FL processes, in order to evaluate the effectiveness and the robustness of the GA.
\section{Simulation Setup} \label{simulation_setup}
\subsection{Network Environment Setup}
\vspace{-4pt}
The wireless communication network, considered in all performed experiments, is comprised of one coordinator and up to 40 ($|\mathcal{K}| \leq 40$) heterogeneous (in terms of resource capabilities) workers. Of all workers, the $20\%$ are of reduced capabilities and are considered as low-end devices, while the rest are considered as high-end. Each low-end device $i$, operates with $f_i^{max} = 1$ GHz, $C_i = 4$ \cite{9475121Mo} and $p_i^{max} = 28$ dBm. Each high-end device $j$, operates with $f_j^{max} = 3$ GHz, $C_j = 2$ and $p_j^{max}=33$ dBm. The $\varsigma_k$ is fixed for all workers and equal to $10^{-28}$ $\mathrm{Watt/Hz^{3}}$ \cite{Yang13037554}. We consider non mobile workers and as such all channel gains can be considered as constant. The channel gain is modeled as $g_k = 127 + 30log_{10}(d_k)$, where $d_k$ is the distance of worker $k$ from the coordinator, randomly selected in the range of $[10,500]$ m and $N_0= -158$ dBm/Hz\cite{9462445Zhou}. All workers are assigned with a fixed bandwidth $b_{k} = 20$ MHz.
\vspace{-6pt}
\subsection{Federated Learning Setup}
\vspace{-4pt}
The FL process considers a Convolutional Neural Network (CNN) for handwritten digit recognition, using the MNIST dataset \cite{lecun2010mnist}. The dataset contains 60K samples in total and each worker is assigned with randomly selected $\mathcal{D}_k$  data samples, with $\mathcal{D}_k \in [800,1200]$. The CNN model comprises of $658,922$ trainable parameters, with size $m = 2.51 $ MB and the complexity of the model is $\alpha = 1,800,348$ FLOPS \cite{keras_flops}. All local training processes should meet the performance target $\eta = 0.5$. The FL process is considered complete, when the global CNN model reaches the performance target $\epsilon0 = 0.04$. The time threshold selected for a synchronized FL process is set to $\mathsf{H} = 13$ sec. 

\begin{figure*}[h!]
\centering
    \begin{subfigure}[b]{0.33\textwidth}
      \centering
      \includegraphics[width=0.98\linewidth]{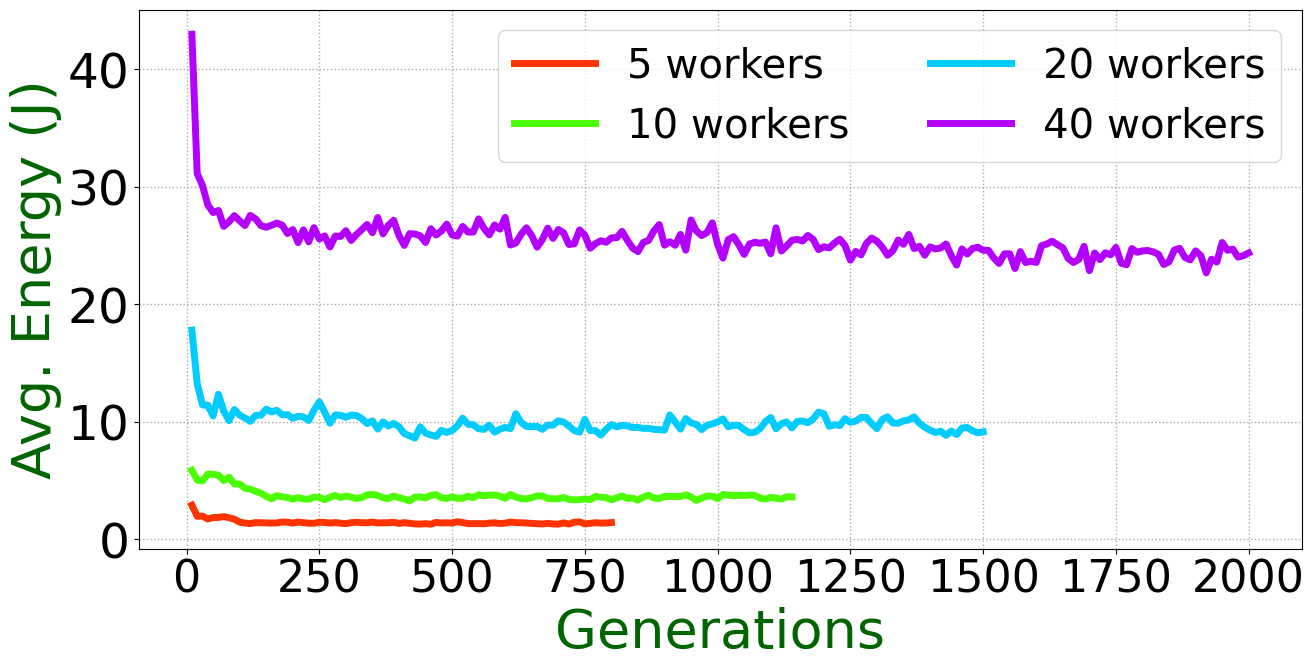}
        \vspace{-7pt}
      \caption{Total Energy Consumption}
      \label{fig:avg_total_energy}
    \end{subfigure}\hfill
    \begin{subfigure}[b]{0.33\textwidth}
      \centering
      \includegraphics[width=0.98\linewidth]{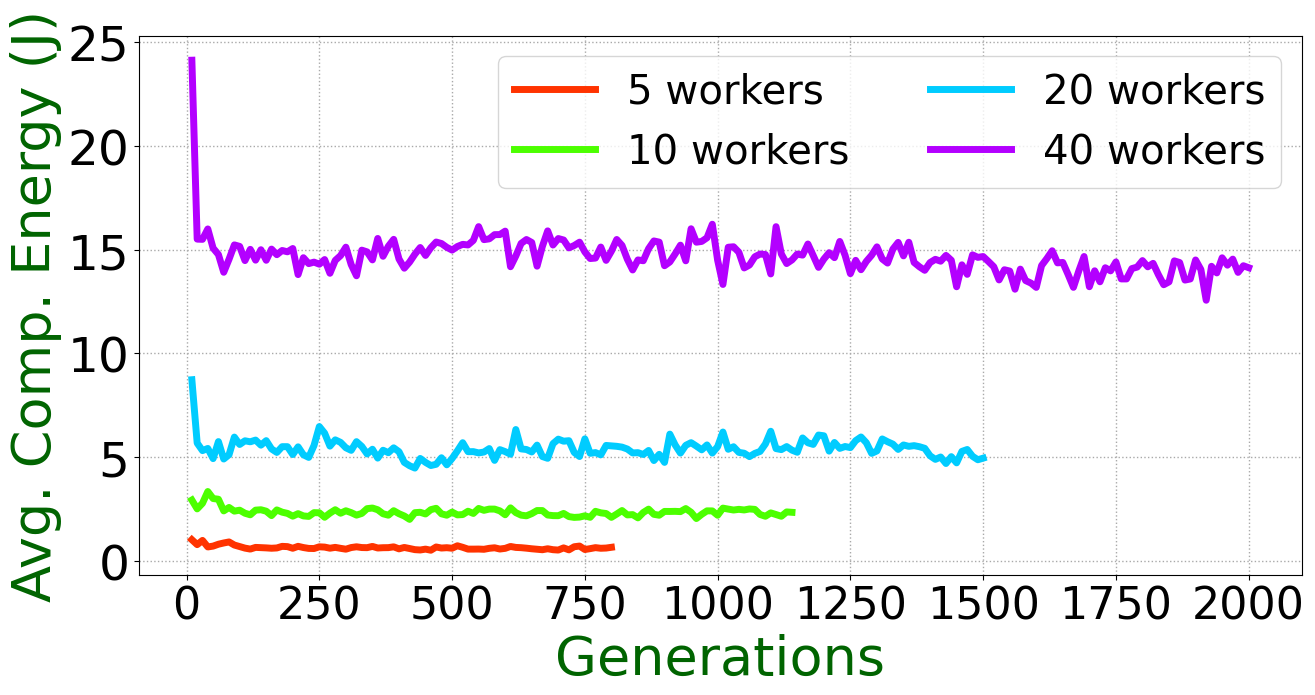}
    \vspace{-7pt}
      \caption{Total Computation Energy}
      \label{fig:avg_comp}
    \end{subfigure}\hfill
    \begin{subfigure}[b]{0.33\textwidth}
      \centering
      \includegraphics[width=0.98\linewidth]{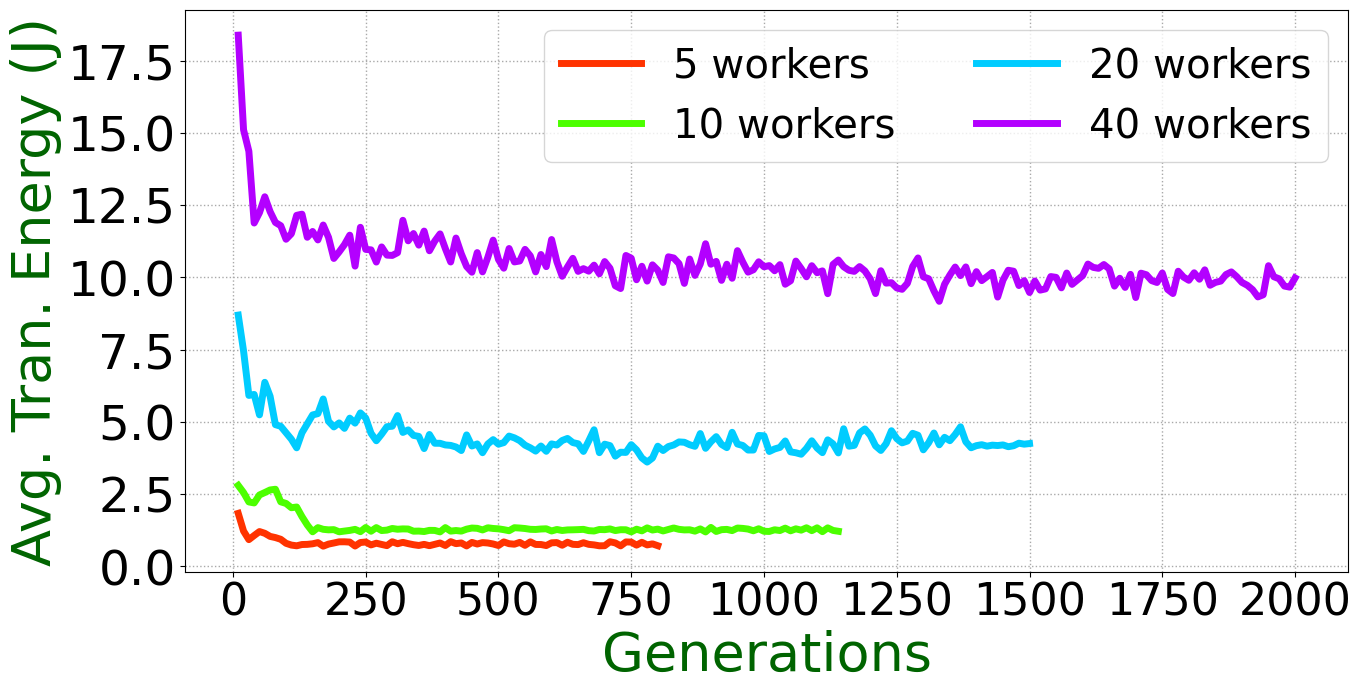}
        \vspace{-7pt}
      \caption{Total Transmission Energy}
      \label{fig:avg_transm}
    \end{subfigure}\hfill
    \begin{subfigure}[b]{0.33\textwidth}
      \centering
      \includegraphics[width=0.98\linewidth]{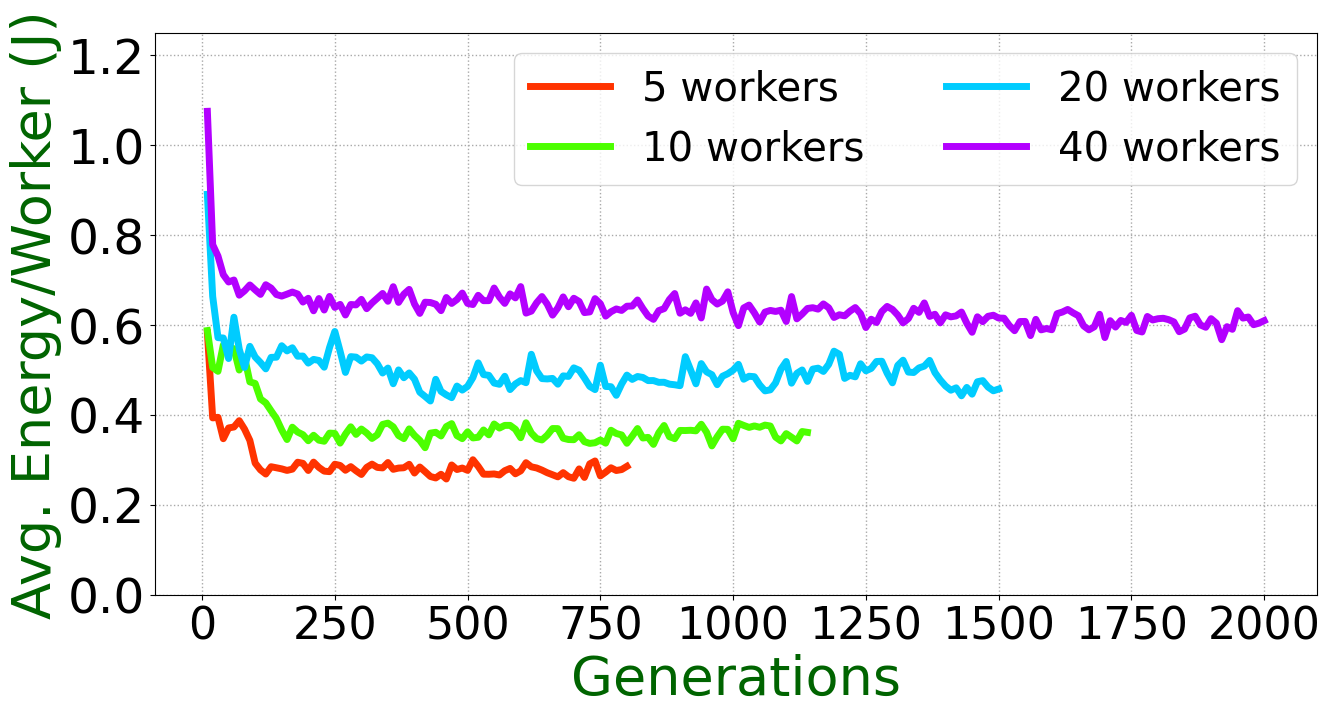}
        \vspace{-7pt}
      \caption{Per Worker Energy Consumption }
      \label{fig:avg_per_worker_energy}
    \end{subfigure}\hfill
    \begin{subfigure}[b]{0.33\textwidth}
      \centering
      \includegraphics[width=0.98\linewidth]{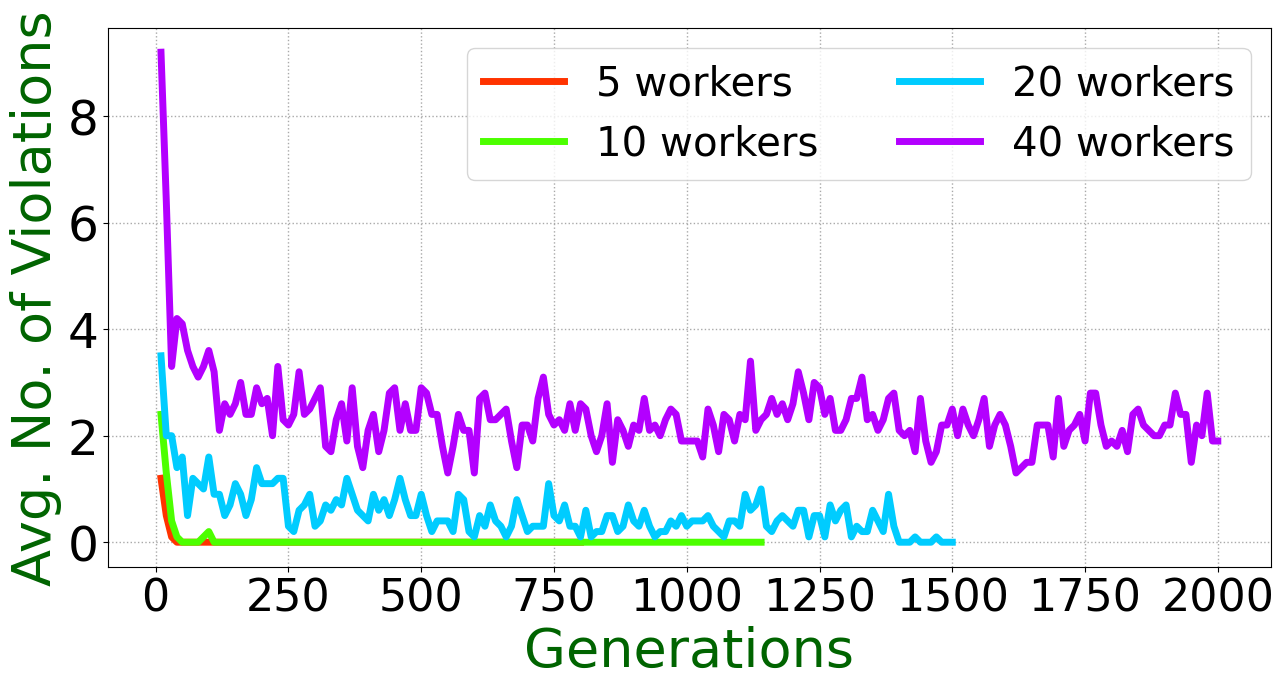}
        \vspace{-7pt}
      \caption{Total Number of Violations}
      \label{fig:avg_violations}
    \end{subfigure}\hfill
    \begin{subfigure}[b]{0.33\textwidth}
      \centering
      \includegraphics[width=0.98\linewidth]{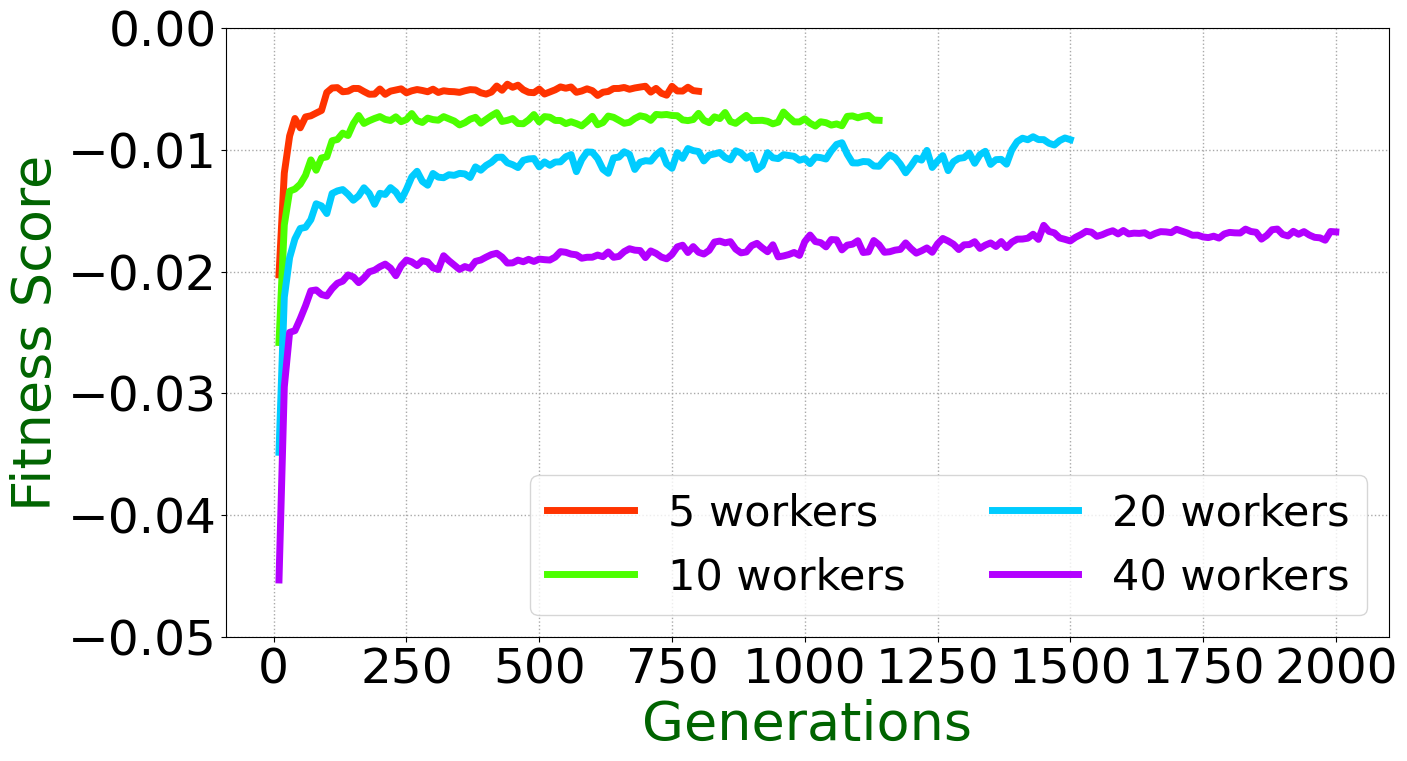}
        \vspace{-7pt}
      \caption{Fitness Score}
      \label{fig:avg_sac_reward}
    \end{subfigure}\hfill
    \caption{Offline phase of GA for different number of workers (5, 10, 20, 40)}\label{fig:train_phase}
    \vspace{-16pt}
\end{figure*}

\vspace{-8pt}
\subsection{Genetic Algorithm Setup}
\vspace{-2pt}

Table \ref{tab:ga_configuration} includes the configuration of the GA. The maximum number of generations for the different number of workers is set to $5000$, while the early stopping is set to $100$ generations. The mutation rate, when the hyper-mutation is triggered, is increased by $50\%$ and the size of the fixed memory is set to $[15, 35, 55, 85]$ for the case of $5, 10, 20$ and $40$ workers, respectively. The hybrid operation is triggered when $\mathsf{D}$ is equal to $[0.4, 0.3, 0.25, 0.2]$ for the four respective cases. 
\begin{table}[h]
\vspace{-7pt}
\centering
\resizebox{160pt}{!}{%
\begin{tabular}{|l|cccc|}
\hline
\textbf{No. of Workers} & \multicolumn{1}{c|}{\textbf{5}} & \multicolumn{1}{c|}{\textbf{10}} & \multicolumn{1}{c|}{\textbf{20}} & \textbf{40} \\ \hline
\textbf{Population Size} & \multicolumn{1}{c|}{40}  & \multicolumn{1}{c|}{120}  & \multicolumn{1}{c|}{210} & 220  \\ \hline
\textbf{Elites}          & \multicolumn{1}{c|}{10}  & \multicolumn{1}{c|}{20}   & \multicolumn{1}{c|}{30}  & 60   \\ \hline
\textbf{Crossover Rate}  & \multicolumn{1}{c|}{0.3} & \multicolumn{1}{c|}{0.3}  & \multicolumn{1}{c|}{0.3} & 0.3  \\ \hline
\textbf{Mutation Rate}   & \multicolumn{1}{c|}{0.1} & \multicolumn{1}{c|}{0.05} & \multicolumn{1}{c|}{0.1} & 0.05 \\ \hline
\end{tabular}%
}
\vspace{-2pt}
\caption{GA Configuration }
\vspace{-10pt}
\label{tab:ga_configuration}
\end{table}

As already mentioned in Section \ref{proposed_solution}, for the offline phase of the GA, a simulated FL environment is used, based on a real FL process. From equation (\ref{objective_function}), it becomes apparent that two FL and energy-related parameters are not known a priori and should be estimated, namely the number of \textit{local} iterations per worker ($I_{k,n}$) and the total number of \textit{global} iterations. Following a statistical analysis of over 100 independent FL executions, the $I_{k,n}$ is selected based on the worker's data samples variance and is in the range $[2, 11]$. Higher data samples variance, means increased data heterogeneity and as a result a higher number of \textit{local} iterations is required to reach $\eta$. According to the statistical analysis, the number of \textit{global} iterations is in the range $[10, 22]$, and it is inversely proportional to the number of \textit{local} iterations. 

\vspace{-2pt}
\section{Performance Evaluation}\label{performance_evaluation}
\vspace{-3pt}
This section provides numerical results to evaluate the
performance of the proposed safe GA based solution. Two main performance metrics are considered in the evaluation phase: \textbf{1)} the total energy consumption of the FL (both computation and communication), \textbf{2)} the convergence speed of the FL (in terms of training time and \textit{global} iterations). 
\vspace{-6pt}
\subsection{Offline Phase}
\vspace{-3pt}
Figure \ref{fig:train_phase} illustrates the performance of the GA, for different number of workers
, as the number of generations increases. Figure \ref{fig:avg_total_energy} depicts the total energy consumption in each generation, that is the sum of the total computation (Fig. \ref{fig:avg_comp}) and transmission energies (Fig. \ref{fig:avg_transm}). As it can be inferred from Figures (\ref{fig:avg_total_energy} - \ref{fig:avg_transm}), the amount of consumed energy, for the different number of workers, significantly decreases with the number of generations. More specifically, the GA achieves an overall approximate reduction in the energy consumption up to 75\%, 76\%, 66\% and 56\% in the case of 5, 10, 20 and 40 workers, respectively. As the number of workers increases, so does the time (in generations) the GA requires to converge. This relies on the fact that the number of genes increases proportionally to the number of workers, and as a result, the GA requires more exploration time. 

In addition, it is intuitive that the higher the number of workers, the higher the total energy consumption of the system. As a result, the fitness score that the GA retrieves is lower for higher number of workers (Fig. \ref{fig:avg_sac_reward}). However, as it is depicted in Figure \ref{fig:avg_per_worker_energy}, the GA achieves an average per worker total energy consumption approximately equal to 0.44 Joules. This behavior showcases the scalability of our solution, since the increase in the total amount of energy consumption is mainly based on the number of workers. 

As already stated, we introduce a penalty function towards reducing wasted resources of the system. Figure \ref{fig:avg_violations} validates our claims, since the total number of constraint violations during an FL process are eliminated. For the most challenging case of 40 workers our solution is still able to provide an efficient strategy reaching in total $1.9$ violations, which is translated to $0.04$ per worker.
\vspace{-7pt}
\subsection{Online Phase}
\vspace{-3pt}
The online phase includes the evaluation of the GA strategy produced by the offline phase. All statistical results are averaged, along with their standard deviation ($\pm$STD), over 100 independent real FL runs. Firstly, a comparison between our proposed solution and two baseline schemes, namely a Random and a Greedy Selection scheduler \cite{9384231Zhan} is performed, for the $5$, $10$, $20$ and $40$ workers.
\begin{itemize}
  \item \textbf{Random Selection Scheduler (RSS): } In each \textit{global} iteration, the RSS orchestrates randomly the resources of each worker, based on their available capacities. 
  \item \textbf{Greedy Selection Scheduler (GSS): } In each \textit{global} iteration, the GSS chooses the resource capacities of all workers that led to the best outcome so far, in terms of the total energy consumption. 
\end{itemize}
\begin{table*}[ht!]
\vspace{2pt}
\centering
\resizebox{490pt}{!}{%
\begin{tabular}{|l|ccc|ccc|ccc|ccc|}
\hline
\multicolumn{1}{|c|}{\textbf{\begin{tabular}[c]{@{}c@{}}Total Avg. \\ ($\pm$STD)\end{tabular}}} &
  \multicolumn{3}{c|}{\textbf{5 Workers}} &
  \multicolumn{3}{c|}{\textbf{10 Workers}} &
  \multicolumn{3}{c|}{\textbf{20 Workers}} &
  \multicolumn{3}{c|}{\textbf{40 Workers}} \\ \hline
\textbf{Schedulers} &
  \multicolumn{1}{c|}{\cellcolor[HTML]{EFEFEF}\textbf{GA}} &
  \multicolumn{1}{c|}{\textbf{RSS}} &
  \textbf{GSS} &
  \multicolumn{1}{c|}{\cellcolor[HTML]{EFEFEF}\textbf{GA}} &
  \multicolumn{1}{c|}{\textbf{RSS}} &
  \textbf{GSS} &
  \multicolumn{1}{c|}{\cellcolor[HTML]{EFEFEF}\textbf{GA}} &
  \multicolumn{1}{c|}{\textbf{RSS}} &
  \textbf{GSS} &
  \multicolumn{1}{c|}{\cellcolor[HTML]{EFEFEF}\textbf{GA}} &
  \multicolumn{1}{c|}{\textbf{RSS}} &
  \textbf{GSS} \\ \hline
\textbf{Total Energy (J)} &
  \multicolumn{1}{c|}{\cellcolor[HTML]{EFEFEF}\begin{tabular}[c]{@{}c@{}}19.3\\ ($\pm$4.5)\end{tabular}} &
  \multicolumn{1}{c|}{\begin{tabular}[c]{@{}c@{}}116.4\\ ($\pm$23.2)\end{tabular}} &
  \begin{tabular}[c]{@{}c@{}}51.5\\ ($\pm$14.6)\end{tabular} &
  \multicolumn{1}{c|}{\cellcolor[HTML]{EFEFEF}\begin{tabular}[c]{@{}c@{}}52.8\\ ($\pm$7)\end{tabular}} &
  \multicolumn{1}{c|}{\begin{tabular}[c]{@{}c@{}}271.1\\ ($\pm$38)\end{tabular}} &
  \begin{tabular}[c]{@{}c@{}}168\\ ($\pm$27.9)\end{tabular} &
  \multicolumn{1}{c|}{\cellcolor[HTML]{EFEFEF}\begin{tabular}[c]{@{}c@{}}136.7\\ ($\pm$13)\end{tabular}} &
  \multicolumn{1}{c|}{\begin{tabular}[c]{@{}c@{}}576.4\\ ($\pm$69.4)\end{tabular}} &
  \begin{tabular}[c]{@{}c@{}}400.7\\ ($\pm$52.3)\end{tabular} &
  \multicolumn{1}{c|}{\cellcolor[HTML]{EFEFEF}\begin{tabular}[c]{@{}c@{}}367.7\\ ($\pm$33.7)\end{tabular}} &
  \multicolumn{1}{c|}{\begin{tabular}[c]{@{}c@{}}1135.7\\ ($\pm$101.2)\end{tabular}} &
  \begin{tabular}[c]{@{}c@{}}850.2\\ ($\pm$85.9)\end{tabular} \\ \hline
\textbf{\begin{tabular}[c]{@{}l@{}}Computation\\ Energy (J)\end{tabular}} &
  \multicolumn{1}{c|}{\cellcolor[HTML]{EFEFEF}\begin{tabular}[c]{@{}c@{}}7.5\\ ($\pm$1.7)\end{tabular}} &
  \multicolumn{1}{c|}{\begin{tabular}[c]{@{}c@{}}51.1\\ ($\pm$12)\end{tabular}} &
  \begin{tabular}[c]{@{}c@{}}25.3\\ ($\pm$7.5)\end{tabular} &
  \multicolumn{1}{c|}{\cellcolor[HTML]{EFEFEF}\begin{tabular}[c]{@{}c@{}}30.5\\ ($\pm$4.4)\end{tabular}} &
  \multicolumn{1}{c|}{\begin{tabular}[c]{@{}c@{}}120.6\\ ($\pm$19.7)\end{tabular}} &
  \begin{tabular}[c]{@{}c@{}}112.2\\ ($\pm$20)\end{tabular} &
  \multicolumn{1}{c|}{\cellcolor[HTML]{EFEFEF}\begin{tabular}[c]{@{}c@{}}61.4\\ ($\pm$6.5)\end{tabular}} &
  \multicolumn{1}{c|}{\begin{tabular}[c]{@{}c@{}}264.4\\ $\pm$35.2)\end{tabular}} &
  \begin{tabular}[c]{@{}c@{}}194.2\\ ($\pm$27.8)\end{tabular} &
  \multicolumn{1}{c|}{\cellcolor[HTML]{EFEFEF}\begin{tabular}[c]{@{}c@{}}166.2\\ ($\pm$17.7)\end{tabular}} &
  \multicolumn{1}{c|}{\begin{tabular}[c]{@{}c@{}}526.9\\ ($\pm$5)\end{tabular}} &
  \begin{tabular}[c]{@{}c@{}}452.9\\ ($\pm$50.7)\end{tabular} \\ \hline
\textbf{\begin{tabular}[c]{@{}l@{}}Transmission\\ Energy (J)\end{tabular}} &
  \multicolumn{1}{c|}{\cellcolor[HTML]{EFEFEF}\begin{tabular}[c]{@{}c@{}}11.8\\ ($\pm$3.2)\end{tabular}} &
  \multicolumn{1}{c|}{\begin{tabular}[c]{@{}c@{}}65.3\\ ($\pm$15.4)\end{tabular}} &
  \begin{tabular}[c]{@{}c@{}}26.2\\ ($\pm$8.4)\end{tabular} &
  \multicolumn{1}{c|}{\cellcolor[HTML]{EFEFEF}\begin{tabular}[c]{@{}c@{}}22.3\\ ($\pm$3.3)\end{tabular}} &
  \multicolumn{1}{c|}{\begin{tabular}[c]{@{}c@{}}150.5\\ ($\pm$24.1)\end{tabular}} &
  \begin{tabular}[c]{@{}c@{}}55.8\\ ($\pm$12.3)\end{tabular} &
  \multicolumn{1}{c|}{\cellcolor[HTML]{EFEFEF}\begin{tabular}[c]{@{}c@{}}75.4\\ ($\pm$8.5)\end{tabular}} &
  \multicolumn{1}{c|}{\begin{tabular}[c]{@{}c@{}}312\\ ($\pm$43.9)\end{tabular}} &
  \begin{tabular}[c]{@{}c@{}}206.5\\ ($\pm$31.9)\end{tabular} &
  \multicolumn{1}{c|}{\cellcolor[HTML]{EFEFEF}\begin{tabular}[c]{@{}c@{}}201.6\\ ($\pm$19.9)\end{tabular}} &
  \multicolumn{1}{c|}{\begin{tabular}[c]{@{}c@{}}608.8\\ ($\pm$59.1)\end{tabular}} &
  \begin{tabular}[c]{@{}c@{}}397.3\\ ($\pm$43.9)\end{tabular} \\ \hline
\textbf{\begin{tabular}[c]{@{}l@{}}Training Time\\ per Global \\ Iteration (s)\end{tabular}} &
  \multicolumn{1}{c|}{\cellcolor[HTML]{EFEFEF}\begin{tabular}[c]{@{}c@{}}8.1\\ ($\pm$2.5)\end{tabular}} &
  \multicolumn{1}{c|}{\begin{tabular}[c]{@{}c@{}}9\\ ($\pm$1.2)\end{tabular}} &
  \begin{tabular}[c]{@{}c@{}}8.9\\ ($\pm$2.1)\end{tabular} &
  \multicolumn{1}{c|}{\cellcolor[HTML]{EFEFEF}\begin{tabular}[c]{@{}c@{}}7.2\\ ($\pm$1.7)\end{tabular}} &
  \multicolumn{1}{c|}{\begin{tabular}[c]{@{}c@{}}11\\ ($\pm$0.8)\end{tabular}} &
  \begin{tabular}[c]{@{}c@{}}7.6\\ ($\pm$1.5)\end{tabular} &
  \multicolumn{1}{c|}{\cellcolor[HTML]{EFEFEF}\begin{tabular}[c]{@{}c@{}}8.8\\ ($\pm$1.6)\end{tabular}} &
  \multicolumn{1}{c|}{\begin{tabular}[c]{@{}c@{}}12.5\\ ($\pm$0.4)\end{tabular}} &
  \begin{tabular}[c]{@{}c@{}}11.9\\ ($\pm$1.1)\end{tabular} &
  \multicolumn{1}{c|}{\cellcolor[HTML]{EFEFEF}\begin{tabular}[c]{@{}c@{}}12\\ ($\pm$1.4)\end{tabular}} &
  \multicolumn{1}{c|}{\begin{tabular}[c]{@{}c@{}}12.9\\ ($\pm$0.1)\end{tabular}} &
  \begin{tabular}[c]{@{}c@{}}12.9\\ ($\pm$0.1)\end{tabular} \\ \hline
\textbf{\begin{tabular}[c]{@{}l@{}}Global Iterations\end{tabular}} &
  \multicolumn{1}{c|}{\cellcolor[HTML]{EFEFEF}\begin{tabular}[c]{@{}c@{}}15.4\\ ($\pm$4)\end{tabular}} &
  \multicolumn{1}{c|}{\begin{tabular}[c]{@{}c@{}}15.6\\ ($\pm$3.5)\end{tabular}} &
  \begin{tabular}[c]{@{}c@{}}15.7\\ ($\pm$4.4)\end{tabular} &
  \multicolumn{1}{c|}{\cellcolor[HTML]{EFEFEF}\begin{tabular}[c]{@{}c@{}}17.6\\ ($\pm$2.6)\end{tabular}} &
  \multicolumn{1}{c|}{\begin{tabular}[c]{@{}c@{}}18.1\\ ($\pm$2.7)\end{tabular}} &
  \begin{tabular}[c]{@{}c@{}}17.4\\ ($\pm$2.8)\end{tabular} &
  \multicolumn{1}{c|}{\cellcolor[HTML]{EFEFEF}\begin{tabular}[c]{@{}c@{}}18\\ ($\pm$2)\end{tabular}} &
  \multicolumn{1}{c|}{\begin{tabular}[c]{@{}c@{}}18.7\\ ($\pm$2.5)\end{tabular}} &
  \begin{tabular}[c]{@{}c@{}}18.4\\ ($\pm$2.5)\end{tabular} &
  \multicolumn{1}{c|}{\cellcolor[HTML]{EFEFEF}\begin{tabular}[c]{@{}c@{}}18.4\\ ($\pm$1.8)\end{tabular}} &
  \multicolumn{1}{c|}{\begin{tabular}[c]{@{}c@{}}18.3\\ ($\pm$1.7)\end{tabular}} &
  \begin{tabular}[c]{@{}c@{}}18.5\\ ($\pm$2)\end{tabular} \\ \hline
\end{tabular}%
}
\caption{Comparison of our solution (GA) against two baseline schedulers (RSS and GSS)}
\vspace{-16pt}
\label{tab:baseline_comparison}
\end{table*}

Table \ref{tab:baseline_comparison} includes the performance comparison of our solution against the baseline schedulers, with regard to: \textbf{1)} the total energy consumption of a complete FL process, \textbf{2)} the training duration of each \textit{global} iteration and \textbf{3)} the total number of \textit{global} iterations. As it can be deduced from Table \ref{tab:baseline_comparison}, our proposed solution outperforms all baseline schedulers, in terms of the energy aspect, while also resulting in the least amount of training time per \textit{global} iteration. More precisely, our solution for the case of 5 workers, achieves a significant reduction in the total energy consumption of $[83, 56]$ \%, compared to the RSS and GSS, respectively. Similarly, for the case of 10, 20 and 40 workers, our solution results in a similar percentage decrease of $[81, 68]$ \%, $[76, 66]$ \% and $[68, 57]$ \%, respectively. As it appears in the total training time per \textit{global} iteration of the FL process, our solution is on average $20\%$ and $12\%$ faster than the RSS and GSS, respectively. Finally, it should be highlighted that the simulated FL environment (offline phase), achieves a similar performance in the total energy consumption (complete FL process), compared to the real FL executions (online phase) for the same number of \textit{global} iterations. More specifically, the distance in the total energy consumption between the offline and online phase, is equal to $2\%$, $8\%$, $8\%$ and $7\%$, in favor of the online phase (lower total energy consumption), for the case of $5$, $10$, $20$ and $40$ workers, respectively, confirming the effectiveness of the simulated FL environment. 

\vspace{-8pt}
\section{Conclusions}\label{conclusions}
\vspace{-4pt}
This paper proposes a safe GA based solution, targeting the minimization of the overall energy consumption of an FL process in a wireless communication network. A simulated FL environment is designed for
the GA's offline phase, targeting lower complexity and faster convergence. 
A penalty function is introduced, towards a safe GA process that almost minimizes the wasted resources of the system. Evaluation results showcase a significant reduction in the overall energy consumption, achieving up to 76\% for the offline phase. For the online phase, a decrease of up to 83\% in the total energy consumption is achieved compared to two state-of-the-art baseline solutions.
\vspace{-3pt}
\bibliographystyle{IEEEtran}
\bibliography{main}

\end{document}